\documentclass[floatfix,notitlepage,aps,nofootinbib,tightenlines,11pt]{revtex4-1}

\usepackage{epsfig}
\usepackage{graphicx}
\usepackage{color}
\usepackage{amsmath}
\usepackage{amssymb}

\usepackage[sort&compress]{natbib}
\newcommand{\myvec}[1]{\ensuremath{\mathbf{#1} } }

\newcommand{\ti}[1]{\ensuremath{\myvec{t}_{#1 } } }
\newcommand{\T}[1]{\ensuremath{\myvec{T}_{#1 } } }

\newcommand{\f}[1]{\ensuremath{\myvec{f}_{#1 } } }

\newcommand{\Lk}{\ensuremath{\mathbf{L}^{\textrm{\scriptsize{-1}}}_{\textrm{k}}}}
\newcommand{\Tau}{\overset{*}{\tau}}
\newcommand{\EXP}[1]{\ensuremath{\big \langle#1 \big \rangle}}
\newcommand{\IB}{\ensuremath{\mathcal{IB }}}
\newcommand{\PIC}{\ensuremath{\mathcal{PIC}}}
\newcommand{\NMAX}{\ensuremath{N_{\textnormal{max}}}}
\newcommand{\PICTOTSR}{\ensuremath{\PIC^{\textnormal{SR}}_{\textnormal{tot}}}}
\newcommand{\PICTOTFB}{\ensuremath{\PIC^{\textnormal{FB}}_{\textnormal{tot}}}}
\newcommand{\PICTOT}{\ensuremath{\PIC_{\textnormal{tot}}}}

\begin{document} 

\title{Optimally fuzzy temporal memory}

\author{Karthik H. Shankar}  
\affiliation{Center for Memory and Brain,  Boston University}
\author{ Marc W. Howard}
\affiliation{Center for Memory and Brain,  Boston University}

\begin{abstract}
 Any learner with the ability to predict the future of a
structured time-varying signal must maintain a memory of  the recent past. If
the signal has a characteristic timescale relevant to future prediction, the
memory can be a simple shift register---a moving window extending into the
past, requiring storage resources that linearly grows  with the timescale to
be represented. However, an independent general purpose learner cannot a
priori know the characteristic prediction-relevant timescale of the signal.
Moreover, many naturally occurring signals show scale-free long range
correlations implying that the natural prediction-relevant timescale is
essentially unbounded. Hence the learner should maintain information from the
longest possible timescale allowed by resource availability. Here we construct
a  fuzzy memory system that optimally sacrifices the temporal accuracy of
information in a scale-free fashion in order to represent prediction-relevant
information from exponentially long timescales.  
Using several illustrative examples, we demonstrate the advantage of the fuzzy
memory system over a shift register in time series forecasting of natural
signals. When the available storage resources are limited, we suggest that a
general purpose learner would be better off committing to such a fuzzy memory
system.  
\end{abstract}

\maketitle{}

\section{Introduction}

Natural learners face a severe computational problem in attempting to predict
the future of time varying signals. Rather than being presented with a large
training set of examples, they must compute on-line using a continuously
evolving representation of the recent past.  A basic question arises here--how
much of the recent past is required to generate future predictions?
Maintaining past information in memory comes with a metabolic cost; 
we would expect a strong evolutionary pressure to minimize the resources
required. A shift register  can accurately represent information
from the recent past up to a chosen timescale, while consuming resources that
grow linearly  with that timescale. However, the prediction-relevant timescale
of the signal is generally unknown prior to learning.  Moreover there are many
examples of naturally occurring signals with scale-free long range
correlations
\citep{VossClar75,Mand82,Fiel87,TorrOliv03,LinkEtal01,Bail96,Gild01,VanEtal03,WageEtal04}, commonly known as $1/f$ signals, making the natural prediction-relevant
timescale essentially unbounded. Our focus is on the following question:  If
an independent general purpose learner is to forecast long range correlated
natural signals, what is the optimal way to represent the past information in
memory with limited resources? 

We argue that the solution is to construct a memory that reflects the natural
scale-free temporal structure associated with the uncertainties of the world.
For example, the timing of an event that happened 100 seconds ago does not
have to be represented as accurately in memory as the timing of an event that
happened 10 seconds ago.  Sacrificing the temporal accuracy of  information in
memory leads to tremendous resource conservation, yielding the capacity to
represent information from exponentially long timescales with 
linearly growing resources.  Moreover, by sacrificing the temporal accuracy of information
in a scale-free fashion, the learner can  gather the relevant statistics from
the signal in a way that is optimal if the signal contains scale-free
fluctuations.  To mechanistically construct such a memory system, it is
imperative to keep in mind that the information represented in memory should
self sufficiently evolve in real time without relying on any information other
than the instantaneous input and what is already represented in memory;
reliance on any external information would require additional storage
resources. In this paper we describe a  fuzzy\footnote{Fuzzy temporal memory is not related to fuzzy logic.}, \emph{i.e.} temporally
inaccurate, memory system that (i)  represents information from very long
timescales, (ii) optimally sacrifices temporal accuracy while maximally
preserving the prediction-relevant information from the past, and (iii)
evolves self sufficiently in real time.

The layout of the paper is as follows. 
In section~2, based on some general properties of long range correlated
signals, we derive the criterion for optimally sacrificing the temporal
accuracy so that the prediction relevant information from exponentially long
time scales is maximally preserved in the memory with finite resources.
However, it is non-trivial to construct such a fuzzy memory  in a self
sufficient way. In section~3, we describe a strategy to construct  a
self sufficient scale-free representation of the recent past. This strategy is
based on a neuro-cognitive model of internal time, TILT \citep{ShanHowa12}, and
is mathematically  equivalent to encoding the Laplace transform of the past
and approximating its inverse to reconstruct a fuzzy representation of the
past. With an optimal choice of a set of memory nodes, this representation
naturally leads to a self-sufficient fuzzy memory system. In section~4, we illustrate the
utility of the fuzzy memory with some simple time series forecasting examples.
Using several very different time series we show that the fuzzy memory
enhances the ability to predict the future in comparison to a shift register
with the same number of nodes.  Of course, representing the recent past in
memory does not by itself guarantee the ability to successfully predict the
future. It is crucial to learn the prediction-relevant statistics underlying
the signal with an efficient learning algorithm. The choice of the learning
algorithm is largely modular to the choice of the memory system.  In this
paper, we sidestep the problem of learning, and only focus on the memory. As a
place holder for a learning algorithm, we use linear regression in the
demonstrations of time series forecasting.

\section{Optimal Accuracy-Capacity tradeoff}

\begin{figure}
\centering
\includegraphics[height=0.3\textheight]{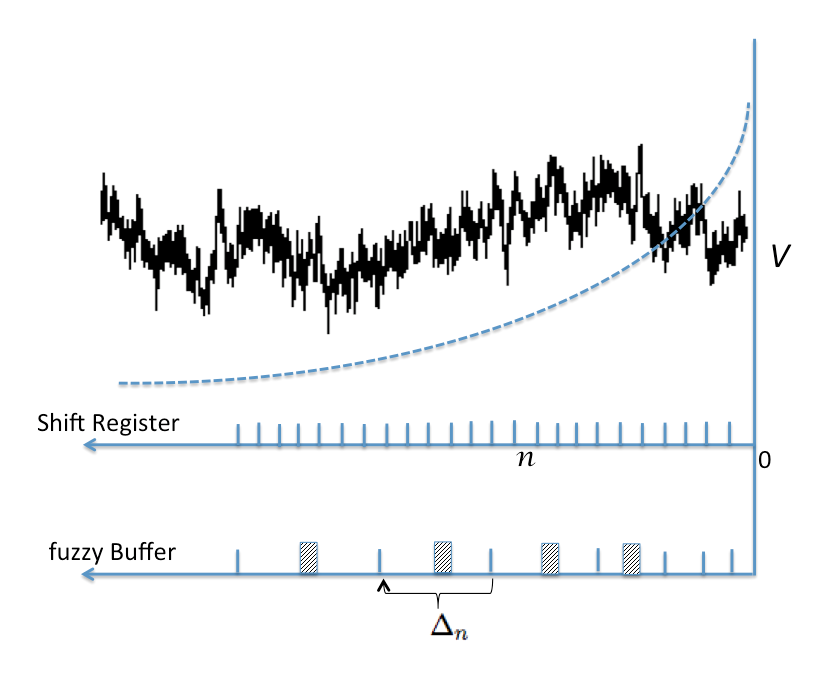}  
\caption{\small A sample time series $V$ with power-law two-point correlation
is plotted w.r.t.~time($n$) with the current time step taken to be $n=0$. The
figure contrasts the way in which the information is represented in a shift
register and the fuzzy buffer. Each node of the fuzzy buffer linearly combines
information from a bin containing multiple shift register nodes. The dashed
curve shows the predictive information content from each time step that is
relevant for predicting the future value of the time series.  \label{fig:SSR}
}
\end{figure}

Suppose a learner needs to learn to predict a real valued time series with
long range correlations. Let $V = \{v_n:  n \in (1,2,3...\infty) \}$ represent
all the past values of  the time series relative to the present moment at $n=0$.  Let
$V$ be a stationary series with zero mean and finite variance. Ignoring the
existence of any higher order correlations, let the two point correlation be
given by $\EXP{v_{n}  v_{m}} \simeq 1/|n-m|^{\alpha}$. When $\alpha \le 1$,
the series will be  long range correlated \citep{Bera94}. The goal of the
learner is to successfully predict the current value $v_o$ at $n=0$.
Fig.~\ref{fig:SSR} shows a sample time series leading up to the
present moment.  The $y$-axis corresponds to the present moment and to its
right lies the unknown future values of the time series.  The $x$-axis labeled
as shift register denotes a memory buffer wherein each $v_n$ is accurately
stored in a unique node.  As we step forward in time, the information stored
in each node  will be transferred to its left-neighbor and the current value
$v_o$ will enter the first node. The shift register can thus self sufficiently
evolve in real time. The longest time scale that can be represented with the shift register is
linearly related to the number of nodes. 

Given a limited number of memory nodes, what is the optimal way to represent
$V$ in the memory so that the information relevant to prediction of $v_o$ is
maximally preserved?  We will show that this is achieved in the fuzzy buffer
shown in fig.~\ref{fig:SSR}. Each node of the fuzzy buffer holds the average
of $v_n$s over a bin.  For the fuzzy buffer, the widths of the bins
increase linearly with the center of the bin; the bins are chosen to tile the
past time line. Clearly, the accuracy of representing $V$ is sacrificed in the
process of compressing the information in an entire bin into a real number,
but note that we attain the capacity to represent information from
exponentially long timescales. With some analysis, we will show that
sacrificing the accuracy with bin widths chosen in this way leads to maximal
preservation of prediction-relevant information.   

It is clear that the fuzzy buffer shown in fig.~\ref{fig:SSR} cannot
evolve self sufficiently in real time; information lost during compression of
a bin at any moment is required at the next moment to recompute the bin
average. At each moment we would explicitly require all the
$v_n$s to correctly update the values in the fuzzy buffer. So the fuzzy
buffer does not actually save resources in representing the time series.
Even though the fuzzy buffer is not self sufficient, it is extremely simple to
analyze; we will use the fuzzy buffer to derive the optimal binning strategy
to maximally preserve prediction-relevant information in a time series with
finite storage resources. The reader who is willing to take the optimality of
linearly-increasing bin widths on faith can skip ahead to the section~3 where
we construct a self-sufficient memory system with that property.  

\subsection{Deriving the optimal binning strategy}

To quantify the prediction relevant information contained in $V$, let us first
review some general properties of the long range correlated series. Since our
aim here is only to represent $V$ in memory and not to actually understand the
generating mechanism underlying $V$, it is sufficient to consider $V$ as being
generated by a generic statistical algorithm, the ARFIMA model
\citep{GranJoye80,Hosk81,WageEtal04}. The basic idea behind the ARFIMA
algorithm is that  white noise at each time step can be fractionally
integrated to generate a time series with long range correlations \footnote{The most general model ARFIMA(p,d,q) can generate time series with both long and short range structures. Here we choose $p=q=0$ to ignore the short range structures.}.
 Hence the time series can be viewed as being generated by an infinite auto-regressive
generating function. In other words, $v_o$ can be generated from the past
series $V$ and an instantaneous Gaussian white noise input $\eta_o$.
\begin{equation}
v_{o} = \eta_o + \sum_{n=1}^{\infty} a(n) v_{n} .
\label{eq:infAR}
\end{equation}
 The ARFIMA algorithm specifies the coefficients  $a(n)$ in terms of the exponent $\alpha$ in the two point correlation function.  
\begin{equation}
a(n) =  \frac{(-1)^{n+1} \Gamma(d+1)}{\Gamma(n+1) \Gamma(d-n+1)},
\label{eq:a1}
\end{equation}
where $d$ is the fractional integration power given by $d=(1-\alpha)/2$. The time series is stationary and long range correlated with finite variance only when $d \in (0,1/2)$ or $\alpha \in (0,1)$ \citep{GranJoye80,Hosk81}.  The asymptotic behavior of $a(n)$ for large $n$ can be obtained by applying Euler's reflection formula and Stirling's formula to approximate the Gamma functions in eq.~\ref{eq:a1}. It turns out that when either $d$ is small or $n$ is large,  
\begin{equation}
a(n)  \simeq  \left[ \frac{\Gamma(d+1) \sin(\pi d)}{\pi} \right] n^{-(1+d)}.
\label{eq:a}
\end{equation}

For the sake of analytic tractability, the following analysis will focus only on small values of $d$. From eq.~\ref{eq:infAR}, note that $a(n)$ is a measure of the relevance of $v_{n}$ in predicting $v_{o}$, which we shall call the $\PIC$--Predictive Information Content of $v_n$. Taylor expansion of eq.~\ref{eq:a} shows that each $a(n)$ is linear in $d$ up to the leading order. Hence in small $d$ limit, any $v_n$ is a stochastic term $\eta_n$ plus a history dependent term of order $\mathcal{O}(d)$.  Restricting to linear order in $d$, it is clear that the total $\PIC$ of $v_n$ and $v_m$ is simply the sum of their individual $\PIC$s, namely $a(n)+a(m)$.  Thus in the small $d$ limit, $\PIC$ is a simple measure of predictive information that also an extensive quantity.  

When the entire $V$ is accurately represented in an infinite shift register, the total $\PIC$ contained in the shift register is the sum of all $a(n)$. This is clearly the maximum attainable value of $\PIC$ in any memory buffer, and it turns out to be 1. 
\begin{equation}
\PIC_{max} = \sum_{n=1}^{\infty} a(n) = 1.
\end{equation}
In a shift register with $\NMAX$ nodes, the total $\PIC$ is
\begin{eqnarray}
\PICTOTSR = \sum_{n=1}^{\NMAX} a(n) &=& 1- \sum_{\NMAX}^{\infty} a(n) \,\, \simeq \,\,
1- \frac{\sin(\pi d) \Gamma(d+1)}{\pi d} \NMAX^{-d} ,  \nonumber \\
&\xrightarrow{\mbox{\tiny{$d \rightarrow 0$}}} & \,\, d \ln \NMAX .
\label{eq:PIC_SR}  
\end{eqnarray}
For any  fixed $\NMAX$, when $d$ is sufficiently small $\PICTOTSR$ will be
very small.  For example, when $\NMAX=100$ and $d=0.1$,
$\PICTOTSR \simeq 0.4$. For smaller values of $d$, observed in many natural
signals, $\PICTOTSR$ would be even lower. When $d=0.01$, for $\NMAX$ as
large as $10000$, the $\PICTOTSR$ is only 0.08. A large portion of the
predictive information lies in long timescales. So the shift register is
ill-suited to represent information from a long range correlated series.     

The $\PICTOT$ for a memory buffer can be increased if each of its nodes
stored a linear combination of many $v_n$s rather than a single $v_n$ as in
the shift register. This can be substantiated through information theoretic
considerations formulated by the Information Bottleneck ($\IB$) method
\citep{TishEtal99}. A multi-dimensional Gaussian variable can be systematically
compressed to lower dimensions by linear transformations while maximizing the
relevant information content \citep{ChecEtal05}. Although $V$  is a
unidimensional time series in our consideration, at any moment the entire $V$
can be considered as an infinite dimensional Gaussian variable since only its
two point correlations are irreducible. Hence it heuristically follows from
$\IB$ that linearly combining the various $v_n$s into a given number of
combinations and representing each of them in separate memory nodes should
maximize the $\PICTOT$. By examining eq.~\ref{eq:infAR}, it is immediately
obvious that if we knew the values of $a(n)$,  the entire $V$ could be
linearly compressed into a single real number $\sum a(n)v_n$  conveying all of
the prediction relevant information.  Unfortunately, this single-node memory
buffer is not self sufficient: at each moment we explicitly need the entire
$V$ to construct the value of the one-node buffer.  
As an unbiased choice that does not require \emph{a priori} knowledge of the
statistics of the time series, we simply consider uniform averaging over a
bin.  Uniform averaging over a bin discards separate information about the
time of the values contributing to the bin.  Given this consideration, the
question is how should we space the bins to maximize the \PIC.

Consider a bin ranging between $n$ and  $n+\Delta_n$. We shall examine the
effect of averaging all the $v_m$s within this bin and representing it in a
single memory node. If all the $v_m$s in the bin are individually represented,
then the $\PIC$ of the bin is $\sum_m a(m)$. Compressing all the $v_m$s into
their average would however lead to an error in prediction of $v_o$; from
eq.~\ref{eq:infAR} this error is directly related to the extent to which the
$a(m)$s within the bin are different from each other. Hence there should be a reduction in the $\PIC$ of the bin. Given the monotonic functional form of $a(n)$, the maximum reduction can only be  $\sum_m |a(m)-a(n)|$.
\footnote{The minimum reduction in $\PIC$ turns out to be $\sum_m |a(m)-a(n+\Delta_n/2)|$. Adopting this value of reduction in $\PIC$ does not change the
the form of eq.~\ref{eq:n_opt}, but increases the value of $c$ and the total $\PIC$ of each bin. }
The net $\PIC$ of the memory node representing the bin average is then   
\begin{eqnarray}
\PIC &=&  \sum_{m=n}^{n+\Delta_n}a(m) \,\,- \sum_{m=n}^{n+ \Delta_n} | a(n) - a(m) |  \\
& \simeq& \frac{n \,a(n)}{d} \left[ 2-2 \left(1+\frac{\Delta_n}{n} \right)^{-d} -  \frac{d \,\Delta_n}{n}\right] \nonumber
\label{eq:PIC_expand}
\end{eqnarray}
The series summation in the above equation is performed by approximating it as
an integral in the large $n$ limit. The bin size that maximizes the $\PIC$ can
be computed by setting the derivative of $\PIC$ w.r.t.~$\Delta_n$ equal to zero. The optimal bin size and the corresponding $\PIC$ turns out to be 
\begin{equation}
\Delta_n^{opt} = \left[ 2^{1/(1+d)}-1\right] \, n, \qquad 
\PIC^{opt}  \simeq \frac{n \,a(n)}{d} \left[ 2+d- (1+d)2^{1/(1+d)}  \right]  .
\label{eq:opt_bin}
\end{equation}

When the total number of nodes  $\NMAX$ is finite, and we want to represent
information from the longest possible timescale, the straightforward choice is
to pick successive bins such that they completely tile up the past time line
as schematically shown by fuzzy buffer in fig.~\ref{fig:SSR}. If we label the
nodes of the fuzzy buffer by $N$, ranging from 1 to $\NMAX$, and denote the
starting point of each bin by $n_N$, then 
\begin{equation}
n_{N+1} = (1+c) n_N
 \qquad \Longrightarrow  \qquad 
n_N = n_1 (1+c)^{(N-1)},
\label{eq:n_opt}
\end{equation}
where $1+c=2^{1/(1+d)} $. Note that the fuzzy buffer can
represent information from timescales of the order $n_{\NMAX}$, which is
exponentially large compared to the timescales represented by a shift register
with $\NMAX$ nodes.  The total $\PIC$ of the fuzzy buffer, $\PICTOTFB$,  can
now be calculated by summing over the $\PIC$s of each of the bins. For
analytic tractability, we focus on small values of $d$ so that $a(n)$ has the
power law form (eq.~\ref{eq:a}) for all $n$. Applying eqns.~\ref{eq:a} and
\ref{eq:opt_bin},   
\begin{equation}
\PICTOTFB  \simeq
 \sum_{N=1}^{\NMAX} \frac{ \sin(\pi d) \Gamma(d+1)}{ \pi d } \left[ 2+d- (1+d)2^{1/(1+d)}  \right]  n_{N}^{-d}
\end{equation}
Taking $n_1= 1$ and $n_N$ given by eq.~\ref{eq:n_opt},  
\begin{eqnarray}
\PICTOTFB &\simeq& \frac{ \sin(\pi d) \Gamma(d+1)}{ \pi d } \left[ 2+d- (1+d)2^{1/(1+d)}  \right]  
 \left[  \frac{1-(1+c)^{-d\ \NMAX}}{1-(1+c)^{-d}} \right]     \nonumber \\
&\xrightarrow{\mbox{\tiny{$d \rightarrow 0$}}} & [\ln4 -1] d\  \NMAX . 
\label{eq:PIC_FB}
\end{eqnarray}

Comparing eqns.~\ref{eq:PIC_SR} and \ref{eq:PIC_FB}, note that when $d$ is
small, the $\PICTOTFB$ of the fuzzy buffer grows linearly with $\NMAX$ while
the $\PICTOTSR$  of the
shift register grows logarithmically with  $\NMAX$. For example, with
$\NMAX=100$ and $d=0.01$,  the $\PICTOTFB$ of the fuzzy buffer is 0.28,
while the $\PICTOTSR$ of the shift register is only 0.045.  Hence when
$\NMAX$ is relatively small, the fuzzy buffer represents a lot more
predictive information than a shift register. 

The above description of the fuzzy buffer corresponds to the ideal case
wherein the neighboring bins do not overlap and uniform averaging is performed
within each bin. Its critical property of linearly increasing bin sizes
ensures that the temporal accuracy of information is sacrificed optimally and
in a scale-free fashion.  However, this ideal fuzzy buffer cannot
self sufficiently evolve in real time because at every moment all $v_n$s are
explicitly needed for its construction. In the next section, we present a
self sufficient memory system that possesses the critical property of the
ideal fuzzy buffer, but  differs from it by having overlapping bins  and
non-uniform weighted averaging within the bins. To the extent the self
sufficient fuzzy memory system resembles the ideal fuzzy buffer, we can expect
it to be useful in representing long range correlated signals in a
resource-limited environment.

\section{Constructing a self sufficient fuzzy memory}

In this section, we first describe a  mathematical basis for representing the recent past
in a scale-free fashion based on a neuro-cognitive model of internal time, TILT \citep{ShanHowa12}. We then describe several critical considerations
necessary to implement this representation of recent past into a discrete set
of  memory nodes.  Like the ideal fuzzy buffer described in the previous
section, this memory representation will sacrifice temporal accuracy to
represent prediction-relevant information over exponential time scales.
But unlike the ideal fuzzy buffer, the resulting memory representation will be self sufficient, without  requiring additional resources  to construct the representation. 
 
\begin{figure}
\begin{center}
\includegraphics[width=0.50\textwidth]{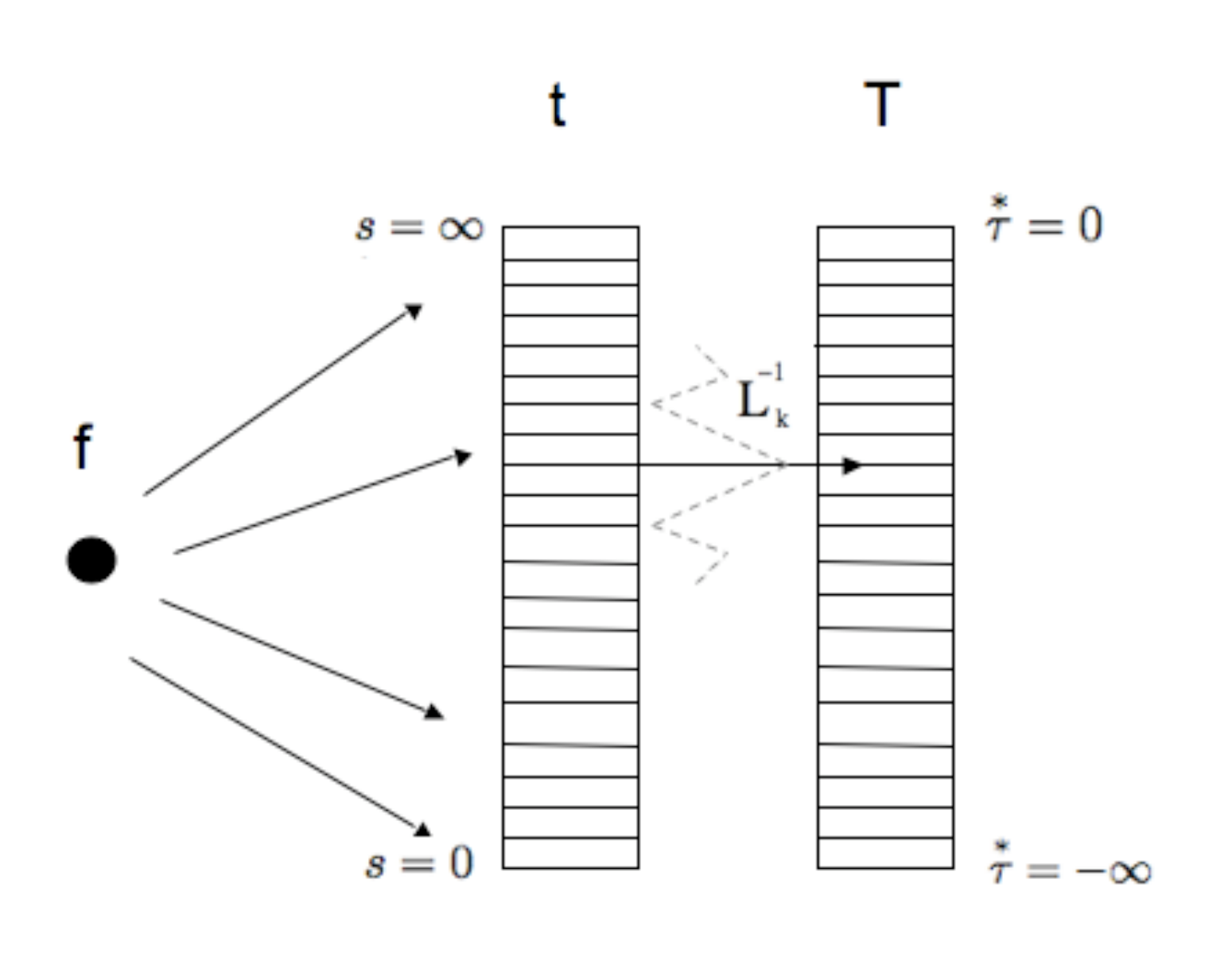}
\end{center}  
\caption{The scale-free fuzzy representation - Each node in the $\ti{}$ column is a
leaky integrator with a specific decay constant $s$ that is driven by the
functional value $\f{}$ at each moment. The activity of the $\ti{}$ column is
transcribed at each moment by the operator $\Lk$  to represent the past
functional values in a scale-free fuzzy fashion in the $\T{}$ column. } 
\label{fig:model}
\end{figure}

Let $\f{}(\tau)$ be a real valued function presented over real time $\tau$.
Our aim now is to construct a memory that represents the past values of
$\f{}(\tau)$ as activity  distributed over a set of nodes with accuracy that
falls off in a scale-free fashion. This is achieved using two columns of nodes
$\ti{}$ and $\T{}$ as shown in fig.~\ref{fig:model}. The $\T{}$ column
estimates \(\f{}(\tau)\) up to the present moment, while the $\ti{}$ column is
an intermediate step used to construct $\T{}$.    The nodes in the $\ti{}$
column are leaky  integrators with decay constants denoted by $s$. Each leaky
integrator independently gets activated by the value of $\f{}$ at any instant
and gradually decays according to 
\begin{equation} \frac{d \, \ti{}(\tau,s)}{d \tau}  = -s \ti{}(\tau,s) + \f{}(\tau). 
\label{eqt} 
\end{equation} 

At every instant, the information in the $\ti{}$ column is transcribed into
the $\T{}$ column through a linear operator $\Lk$.  
\begin{eqnarray}
\T{}(\tau,\Tau) &=&  \frac{(-1)^{k}}{k!}  s^{k+1} \ti{}^{(k)}(\tau,s) \,\, : \,\, \textrm{where}   \,\, \,s=-k/\Tau
 \label{eqT}
 \\ 
\T{} &\equiv& \Lk \big[ \ti{} \big]  \nonumber.
\end{eqnarray}
 Here $k$ is a positive integer and $\ti{}^{(k)}(\tau,s)$ is the $k$-th
derivative of $\ti{}(\tau,s)$ with respect to $s$. The nodes of the $\T{}$
column are labeled by the parameter $\Tau$ and are in  one to one
correspondence with the nodes of the $\ti{}$ column labeled by $s$. The
correspondence between $s$ and $\Tau$ is  given by $s=-k/\Tau $. We refer to
$\Tau$ as \emph{internal time};  at any moment $\tau$, a $\Tau$ node
estimates the value of $\f{}$ at a time $\tau+\Tau$ in the
past. The range of values of  $s$ and $\Tau$ can be made as large as needed at
the cost of resources, but for mathematical idealization we
let them have an infinite range.  

\begin{figure}
\begin{center}
\includegraphics[width=0.6\columnwidth]{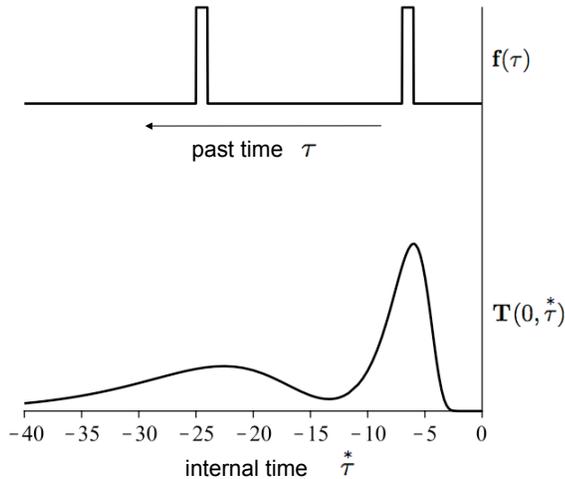}
\end{center}
\caption{Taking the present moment to be $\tau=0$, a sample $\f{}(\tau)$ is plotted in the top, and  the momentary activity distributed across the $\T{}$ column nodes is plotted in the bottom.  }
\label{fig:multitrace}
\end{figure}

The mathematical inspiration of this approach comes from the fact that
$\ti{}(\tau,s)$ encodes the Laplace transform of the entire
history of the function $\f{}$ up to time $\tau$, and the operator $\Lk$
approximately  inverts the Laplace transform \citep{Post30}.
As $k
\rightarrow \infty$,  $\T{}(\tau,\Tau)$ becomes a faithful reconstruction of the
history of $\f{}$ from $-\infty$ to $\tau$,  that is $\T{}(\tau, \Tau) \simeq
\f{}(\tau+\Tau)$ for all values of $\Tau$ from 0 to $-\infty$.  When
$k$ is finite $\T{}(\tau, \Tau)$ is an inaccurate reconstruction of the
history of $\f{}$.  For example, taking the current moment to be $\tau=0$,
fig.~\ref{fig:multitrace} illustrates an $\f{}$ that is briefly non-zero
around  $\tau=-7$ and $\tau=-23$. The reconstructed history of $\f{}$ in the
$\T{}$ column shows two peaks approximately around $\Tau=-7$ and $\Tau=-23$ . 
The value of $\f{}$ at any particular moment in the past is thus smeared over
a range of $\Tau$ values, and this range of smearing increases as we go deeper
into the past. Thus, the more distant past is reconstructed with a lesser
temporal accuracy.   
 
Furthermore, it turns out that the smear is precisely scale invariant. To illustrate this, consider $\f{}(\tau)$ to be a Dirac delta function at a moment $\tau_o$ in the past, $\f{}(\tau) = \delta(\tau-\tau_o)$, and let the
present moment be $\tau=0$.  Applying eqns.~\ref{eqt} and \ref{eqT}, we obtain
\begin{equation}
\T{}(0,\Tau) = \frac{1}{|\tau_o|} \frac{k^{k+1}}{k!} \left( \frac{\tau_o}{\Tau }\right)^{k+1} e^{-k (\tau_o/\Tau)}
\label{eq:deltafunction}
\end{equation}
In the above equation both $\tau_o$ and $\Tau$ are negative. $\T{}(0,\Tau)$ is
the fuzzy reconstruction of the delta function input. $\T{}(0,\Tau)$ is a
smooth peaked function whose height is proportional to $1/|\tau_o|$, width is
proportional to $|\tau_o|$, and the area is equal to 1. Its dependence on the
ratio
$(\tau_o/\Tau)$ ensures scale invariance--for any  $\tau_o$ we can linearly
scale the $\Tau$ values to hold the shape of the function fixed.  In this
sense, $\T{}$ represents the history of $\f{}$ with a scale invariant smear.
To quantify how much smear is introduced, we can estimate the width  of the
peak as the standard deviation $\sigma$ of $\T{}(0,\Tau)$ from the above
equation, which for $k>2$ turns out to be 
\begin{equation}
\sigma[ \T{}(0,\Tau)] = \frac{|\tau_o|}{\sqrt{k-2}} \left[\frac{k}{k-1}\right] .
\label{eq:smear}
\end{equation} 
 Note that $k$ is the only free parameter that affects the smear; 
$k$ indexes the smear in the representation. The larger the
$k$ the smaller the smear. In the limit
$k \rightarrow \infty$, the smear vanishes and the delta function input
propagates into the $\T{}$ column  exactly as delta function without
spreading, as if $\T{}$ were a shift register.

Though we took $\f{}$ to be a simple delta function input to illustrate the scale invariance of the $\T{}$ representation, we can easily evaluate the $\T{}$ representation of an arbitrary $\f{}$ from eqns.~\ref{eqt} and \ref{eqT}.
\begin{equation}
\T{}(0,\Tau) = \int_{-\infty}^{0} \left[ \frac{1}{|\Tau|} \frac{k^{k+1}}{k!} \left( \frac{\tau'}{\Tau }\right)^{k} e^{-k (\tau'/\Tau)} \right]  \f{}(\tau') \, d\tau'
\label{eq:total_T}
\end{equation} 

The $\f{}$ values from a range of past times are linearly combined and
represented in each $\Tau$ node. The term in the square brackets in the above
equation is the weighting function of the linear combination. Note that it is
not a constant function over a circumscribed past time bin,  rather it is a
smooth function peaked at $\tau' = \Tau$, with a spread proportional to
$|\Tau|$. Except for the fact that this weighting function is not uniform, the
activity of a $\Tau$ node has the  desired property mimicking  a bin of the
ideal fuzzy buffer described in section~2.  

\subsection{Self Sufficient discretized implementation}

Although eq.~\ref{eq:total_T} sums over all values of past time, $\T{}$ is
nonetheless a self-sufficient memory representation.  Since eq.~\ref{eqt} is a
local differential equation, the activity of each $\ti{}$ node will
independently evolve in real time, depending only on the present value of the
node and the input available at that moment. Hence any discrete set of $\ti{}$
nodes also evolves self sufficiently in real time. To the extent the activity in a discrete set of $\T{}$ nodes can be constructed from a discrete set of $\ti{}$ nodes, this memory system as a whole can self sufficiently evolve in real time.   
Since the activity of each $\Tau$ node in the $\T{}$ column is constructed
independently of the activity of other $\Tau$ nodes, we can choose any discrete set of $\Tau$
values to form our memory system. In accordance with our analysis of the ideal fuzzy buffer in section~2, we
shall pick the following nodes.  
\begin{equation}
\Tau_{min} \,, \,\,\Tau_{min}(1+c) \,\,,\,\,\Tau_{min}(1+c)^2 \,\,,  \ldots
\,\,  \,\,\, \Tau_{min}(1+c)^{(N_{\textnormal{max}}-1)} =\Tau_{max}.
\label{eq:Tlist}
\end{equation}
Together, these nodes form the fuzzy memory representation with $\NMAX$ nodes.
The spacing in eq.~\ref{eq:Tlist} will yield several important properties.  

Unlike the ideal fuzzy buffer described in section 2, it is not possible to
associate a circumscribed bin  for a node because the weighting function (see
eq.~\ref{eq:total_T}) does not have compact support. Since the weighting
function associated with neighboring nodes overlap with each other, it is
convenient to view the bins associated with neighboring nodes as partially
overlapping. The overlap between neighboring bins implies that some
information is redundantly represented in multiple nodes.  We shall show in
the forthcoming subsections that by appropriately tuning the parameters $k$
and $c$, this information redundancy can be minimized and equally spread in a
scale-free fashion. 

\subsubsection{Discretized derivative} 

Although any set of $\Tau$ nodes could be picked to form the memory buffer,
their activity is ultimately constructed from the nodes in the $\ti{}$ column
whose $s$ values are  given by the one to one correspondence $s=-k/\Tau$.
Since the $\Lk$ operator has to take the $k$-th derivative along the $s$ axis,
$\Lk$ depends on the way the $s$-axis is discretized.

For any discrete set of $s$ values, we can define a linear operator that
implements a discretized derivative.   For notational convenience, let us
denote the activity at any moment $\ti{}(\tau,s)$ as simply $\ti{}(s)$. Since
$\ti{}$ is a column vector with the rows labeled by $s$, we can construct a
derivative matrix $[\mathrm{D}]$ such that
\begin{equation}
\ti{}^{(1)} = [\mathrm{D}] \ti{}  \qquad \Longrightarrow \qquad \ti{}^{(k)} = [\mathrm{D}]^{k} \ti{}     
\end{equation}
The individual elements in the square matrix $[\mathrm{D}]$ depends on the set of $s$ values. To compute these elements, consider any three successive nodes with $s$ values $s_{-1}, s_{o}, s_{1} $. The discretized first derivative of $\ti{}$ at $s_{o}$ is given by 
\begin{equation}
\ti{}^{(1)}(s_o) = \frac{\ti{}(s_{1}) - \ti{}(s_{o})}{s_{1}- s_{o}} \left[\frac{s_{o}-s_{-1}}{s_{1}-s_{-1}}\right]   +\frac{\ti{}(s_{o}) - \ti{}(s_{-1})}{s_{o}- s_{-1}} \left[ \frac{s_{1}-s_{o}}{s_{1}-s_{-1}} \right] 
\end{equation}   
The row in $[\mathrm{D}]$ corresponding to $s_{o}$ will have non-zero entries
only in the columns corresponding to $s_{-1}$, $s_{o}$ and  $s_{1}$. These
three entries can be read out as coefficients of $\ti{}(s_{-1})$,
$\ti{}(s_{o})$ and $\ti{}(s_{1})$ respectively in the r.h.s of the above
equation. Thus the entire matrix $[\mathrm{D}]$ can be constructed from any
chosen set of $s$ values. By taking the $k$-th power of $[\mathrm{D}]$, the
$\Lk$ operator can be straightforwardly  constructed and the activity of the
chosen set of $\Tau$ nodes  can be calculated at each moment.\footnote{Note
that we need $k$ extra nodes in the top and bottom of the $\ti{}$ column in
addition to those that come from one to one correspondence with the chosen
$\Tau$ values.}  This memory system can thus self sufficiently evolve in real
time.

When the spacing between the nodes (controlled by the parameter $c$) is small,
the discretized $k$-th derivative will be accurate. Under uniform
discretization of the $s$ axis, it can be shown that the relative error in
computation of the $k$-th derivative due to discretization is of the order
$\mathcal{O}(k\delta_s^2/24)$, where $\delta_s$ is the distance between
neighboring $s$ values (see appendix B of \citep{ShanHowa12}). Based on the $s$
values corresponding to eq.~\ref{eq:Tlist}, it turns out that the relative
error in the construction of the activity of a $\Tau$ node is
$\mathcal{O}(k^3 c^2 /96 \Tau^2)$.  For large $\Tau$ the error is quite small
but for small $\Tau$ the error can be significant. To curtail the
discretization error, we need to hold $\Tau_{min}$ sufficiently far from zero.
The error can also be controlled by choosing small $c$ for large $k$ and vice
versa. If for practical purposes we require a very small $\Tau_{min}$, then ad
hoc strategies can be adopted to control the error at low $\Tau$ nodes. For
example, by relaxing the requirement of one to one correspondence between the $\ti{}$ and $\T{}$ nodes, we can choose a separate set of closely spaced $s$ values to
exclusively compute the activity of each of the small $\Tau$ nodes. 

Finally, it has to be noted that the discretization error induced in this
process should not be considered as an error in the conventional sense of
numerical solutions to differential equations.  While numerically evolving
differential equations with time-varying boundary conditions, the
discretization error in the derivatives will propagate leading to large errors
as we move farther away from the boundary at late times. But in the situation
at hand, since the activity of each $\Tau$ node is computed independently of
others, the discretization error does not propagate.  Moreover, it should be
noted that the effect of discretization can be better viewed as
coarse-graining the $k$-th derivative rather than as inducing an error in
computing the $k$-th derivative. The fact that each $\Tau$ node ultimately
holds a scale-free coarse-grained value of the input function (see
eq.~\ref{eq:total_T}), suggests that we wouldn't need the exact $k$-th
derivative to construct its activity. To the extent the discretization is
scale-free as in eq.~\ref{eq:Tlist}, the $\T{}$ representation constructed
from the coarse-grained $k$-th derivative will represent some scale free
coarse grained value of the input; however the weighting function would not
exactly match that in eq.~\ref{eq:total_T}.  In other words, even if the
discretized implementation does not accurately match the continuum limit, it
still accurately satisfies the basic properties we require from a fuzzy memory
system.  

\subsubsection{Signal-to-noise ratio with optimally-spaced nodes}

The linearity of equations \ref{eqt} and \ref{eqT} implies that any noise in
the input function $\f{}(\tau)$ will exactly be represented in $\T{}$ without
any amplification. However when there is random uncorrelated noise in the
$\ti{}$ nodes, the discretized $\Lk$ operator can amplify that noise, more so
if $c$ is very small.  It turns out that choice of nodes according to eq.~\ref{eq:Tlist}
results in a constant signal-to-noise ratio across time scales. 

If uncorrelated noise with standard
deviation $\eta$ is added to the activation of each of the $\ti{}$ nodes, then the
$\Lk$ operator combines the noise from $2k$ neighboring nodes leading to a
noisy $\T{}$ representation. If the spacing between the nodes neighboring a
$s$ node is $\delta_s$, then the standard deviation of the noise generated by the $\Lk$
operator is approximately $\eta \sqrt{2^{k}}s^{k+1}/\delta _{s}^{k} k!$. To
view this noise in an appropriate context, we can compare it to the magnitude
of the representation of a delta function signal at a past time $\tau_o=-k/s$
(see eq.~\ref{eq:deltafunction}). The magnitude of the $\T{}$ representation
for a delta function signal is approximately $k^k e^{-k}s/k! $. The signal to
noise ratio (SNR) for a delta function signal is then 
\begin{equation}
\mathrm{SNR}= \eta^{-1} \left( \frac{k [\delta_{s}/s]}{\sqrt{2} e }  \right)^k 
\label{eq:SNR}
\end{equation}
If the spacing between neighboring nodes changes such that  $[\delta_{s} / s]$
remains a constant for all $s$, then the signal to noise ratio will remain
constant over all timescales. This would however require that the nodes are
picked according to eq.~\ref{eq:Tlist}, making SNR = $\eta^{-1}
(kc/\sqrt{2}e)^k$. Any other set of nodes  would make the signal to noise
ratio zero either at large or small timescales. 

This calculation however does not represent the most realistic situation.  Because the
$\ti{}$ nodes are  leaky integrators (see eq.~\ref{eqt}), 
the white noise present across time will accumulate; nodes with long time
constants should have a higher value of $\eta$.  In fact, the standard
deviation of white noise in the $\ti{}$ nodes should go down  with $s$ according to $\eta \propto 1/\sqrt{s}$.
Substituting into Eq.~\ref{eq:SNR}, this would suggest that the SNR of large $\Tau$ nodes
should drop down to zero as $1/\sqrt{|\Tau|}$.\footnote{Interestingly, the
SNR of a shift register would have
a similar property if white noise is introduced in each node of the chain. }

However, because each $\Tau$ node represents  a weighted temporal average of the past signal, it is not appropriate to use an isolated delta function signal to estimate the signal to noise ratio.
It is more appropriate to compare temporally averaged noise to a temporally
spread signal. We consider two such signals. (i) Suppose $\f{}(\tau)$ itself is a temporally uncorrelated white noise like signal. The standard deviation in the activity of a $\Tau$
node in response to this signal is proportional to $1/\sqrt{|\Tau|}$ (see
eq.~\ref{eqwhiteVar} in the appendix). The SNR for this temporally-extended
signal is a constant over all $\Tau$ nodes.  (ii) Consider a purely
positive signal where $\f{}(\tau)$ is a sequence of delta function spikes generated by a Poisson
process. The total expected number of spikes that would be generated in the
timescale of integration of  a $\Tau$ node is simply $\sqrt{|\Tau|}$.
Consequently, the expectation value of the activity of  a $\Tau$ node in
response to this signal would be $\sqrt{|\Tau|}$ multiplied by the magnitude
of representation of a single delta function. The SNR for this signal is
again a constant over all $\Tau$ nodes.  So, we conclude that for any
realistic stationary signal spread out in time, the SNR will be a constant
over all timescales as long as the nodes are chosen according to
eq.~\ref{eq:Tlist}.

\subsection{Information Redundancy}
  
The fact that a  delta function input in $\f{}$ is smeared over many $\Tau$
nodes implies that there is a redundancy in information representation in the
$\T{}$ column.  In a truly scale-free memory buffer, the redundancy in
information representation should be equally spread over all time scales
represented in the buffer. The information redundancy can be quantified in
terms of the mutual information shared between neighboring nodes in the
buffer. In the appendix, it is shown that  in the presence of scale free input
signals, the mutual information shared by any two neighboring buffer nodes can
be a constant only if the $\Tau$ nodes are distributed according to
eq.~\ref{eq:Tlist}. Consequently information redundancy is uniformly
spread only when the $\Tau$ nodes are given by eq.~\ref{eq:Tlist}.

\begin{figure}
\begin{center}
\includegraphics[height=0.15\textheight]{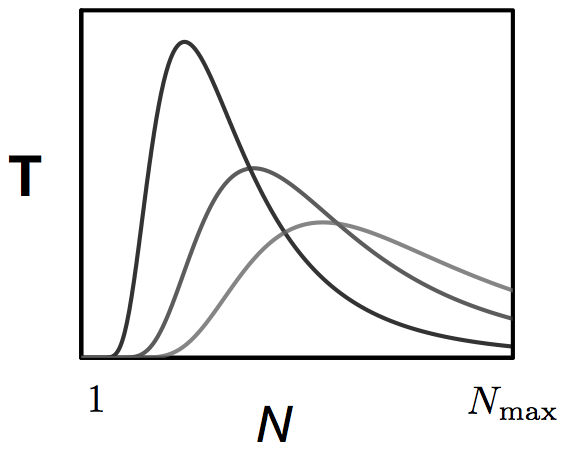} \qquad \qquad \qquad
\includegraphics[height=0.15\textheight]{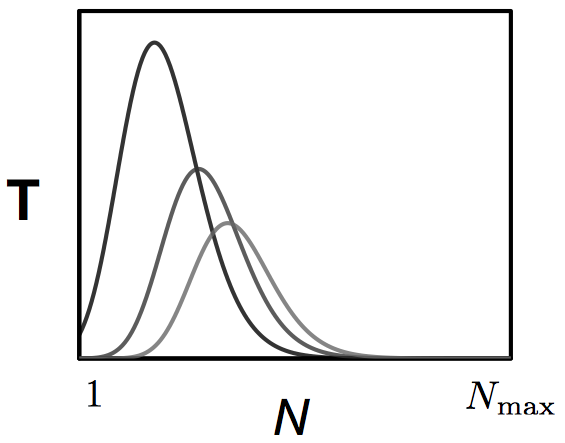}
\end{center}
\caption{Activity of the fuzzy memory nodes following a delta function input
at  three different times is plotted. (Left) the $\Tau$ values of the nodes
are uniformly spaced. (Right) the $\Tau$ values conform to eq.~\ref{eq:Tlist}.
}
\label{fig:InfRed}
\end{figure}

The uniform spread of information redundancy can be intuitively understood by
analyzing how a delta function input spreads through the buffer nodes as time
progresses. Fig.~\ref{fig:InfRed} shows the activity of the buffer nodes at
three points in time following the input. In the left panel, the $\Tau$ values
of the buffer nodes are chosen to be equidistant. Note that as time
progresses, the activity is smeared over many different nodes, implying that
the information redundancy is large in the nodes representing long timescales.
In the right panel, the $\Tau$ values of the buffer nodes are chosen according
to eq.~\ref{eq:Tlist}. Note that as time progresses the activity pattern does
not smear, instead the activity as a whole gets translated with an overall
reduction in size.  The translational invariance of the activity pattern as it
passes through the buffer nodes explains why the information redundancy
between any two neighboring nodes is a constant.  The translational invariance
of the activity pattern can be analytically established as follows. 

Consider  two different values of $\tau_o$ in eq.~\ref{eq:deltafunction}, say $\tau_1$ and $\tau_2$. Let the corresponding $\T{}$ activities be $\T{1}(0,\Tau)$ and $\T{2}(0,\Tau)$  respectively. If the $\Tau$ value of the $N$-th node in the buffer is given by $\Tau_{N}$, then the pattern of activity across the nodes is translationally invariant if and only if $\T{1}(0,\Tau_{N}) \propto \T{2}(0,\Tau_{N+m})$ for some constant integer $m$.  For this to hold true, we need the quantity 
\begin{equation}
\frac{\T{1}(0,\Tau_{N})}{\T{2}(0,\Tau_{N+m})} =  \left( \frac{\tau_1}{\tau_2} \right)^k 
\left[  \frac{\Tau_{N+m}}{\Tau_{N}}    \right]^{k+1} e^{k\left[  \frac{\tau_1}{\Tau_N}-\frac{\tau_2}{\Tau_{N+m}}  \right] }
\end{equation}
to be independent of $N$. This is possible only when the quantity inside the power law form and the exponential form are separately independent of $N$. The power law form can be independent of $N$ only if $\Tau_{N} \propto (1+c)^{N}$, which implies the buffer nodes have $\Tau$ values given by eq.~\ref{eq:Tlist}. The exponential form  is generally dependent on $N$ except when its argument is zero, which happens whenever  $(1+c)^m=\tau_2/\tau_1$ for some integer $m$. For any given $\tau_1$, there are infinitely many $\tau_2$ values for which the condition holds. Moreover when $c$ is small, the condition will approximately hold for any $\tau_2$. Hence, the translational invariance of the activity pattern holds only when the $\Tau$ values of the buffer nodes conform to eq.~\ref{eq:Tlist}.

\subsection{Balancing information redundancy and information loss}

We have seen that the choice of $\Tau$ nodes in accordance with
eq.~\ref{eq:Tlist}  ensures that the  information redundancy is equally
distributed over the buffer nodes. However, equal distribution of information
redundancy is not sufficient; we would also like to minimize information
redundancy.  It turns out that we cannot arbitrarily reduce the information
redundancy without creating information loss. The parameters $k$ and $c$ have
to be tuned in order to balance information redundancy and information loss.
If $k$ is too small for a given $c$, then many nodes in the buffer will
respond to input from any  given moment in the past, resulting in information
redundancy.  On the other hand, if $k$ is too large for a given $c$, the
information from many moments in the past will be left unrepresented in any of
the buffer nodes, resulting in information loss.  So we need to match $c$ with
$k$ to simultaneously  minimize both information redundancy and information
loss. 

This can be achieved if  the information from any given moment in the past is
not distributed over more than two neighboring buffer nodes. To formalize
this, consider a delta function input at a time $\tau_o$ in the past and let
the current moment be $\tau=0$. Let us  look at the activity induced by this
input (eq.~\ref{eq:deltafunction}) in four successive buffer nodes, $N-1$, $N$
and $N+1$ and $N+2$. The $\Tau$ values of these nodes are given by
eq.~\ref{eq:Tlist}, for instance $\Tau_{N}= \Tau_{min} (1+c)^{N-1}$ and
$\Tau_{N+1}= \Tau_{min} (1+c)^{N}$. From eq.~\ref{eq:deltafunction}, it can be
seen that  the $N$-th node attains its maximum activity when $\tau_o=\Tau_N$
and the $(N+1)$-th node attains its maximum activity when $\tau_o=\Tau_{N+1}$,
and for all the intervening times of $\tau_o$ between $\Tau_N$ and
$\Tau_{N+1}$, the information about the delta function input will be spread
over both $N$-th and the $(N+1)$-th nodes. To minimize the information
redundancy, we simply require that when $\tau_o$ is in between $\Tau_{N}$ and
$\Tau_{N+1}$, all the  nodes other than the $N$-th and the $(N+1)$-th nodes
should have almost zero activity. 

\begin{figure}
\begin{center} 
\begin{tabular}{lclc}
\textbf{a} && \textbf{b}\\
&\includegraphics[height=0.25\textheight]{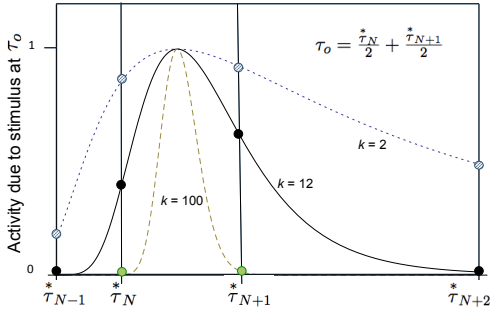}
&&\includegraphics[height=0.25\textheight]{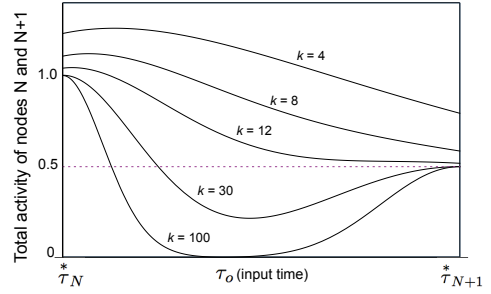}
\end{tabular}
\end{center}

\caption{\textbf{a}. The activity of four successive nodes with $\Tau$ values
given by $\Tau_{N-1}$, $\Tau_{N}$, $\Tau_{N+1}$, and $\Tau_{N+2}$ 
in response to a delta function input at a past moment $\tau_o = (\Tau_{N}/2 +
\Tau_{N+1})/2$. The nodes are chosen according to the distribution given by
eq.~\ref{eq:Tlist} with $c=1$. \textbf{b}. The sum of activity of the nodes
$\Tau_{N}$ and $\Tau_{N+1}$ in response to a delta function input at various
times $\tau_o$ ranging between $\Tau_{N}$ and $\Tau_{N+1}$. For each $k$, the
activities are normalized  to have values in the range of 0 to 1. }
\label{fig:optimalk}
\end{figure}

Fig.~\ref{fig:optimalk}a plots the activity of the four successive nodes with
$c=1$, when $\tau_o$ is exactly in the middle of $\Tau_{N}$ and $\Tau_{N+1}$.
For each value of $k$, the activity is normalized so that it lies between 0
and 1. The four vertical lines represent the 4 nodes and the dots represent
the activity of the corresponding nodes. Note that for $k=2$ the activity of
all 4 nodes is substantially different from zero, implying a significant
information redundancy. At the other extreme, $k=100$, the activity of all the
nodes are almost zero, implying that the information about the delta function
input at time $\tau_o = (\Tau_{N} +\Tau_{N+1})/2$ has been lost.  To minimize
both the information loss and the information redundancy, the value of $k$
should be neither too large nor too small. Note that for $k=12$, the
activities of the $(N-1)$-th and the $(N+2)$-th nodes are almost zero, but
activities of the $N$-th and $(N+1)$-th nodes are non-zero. 

For any given $c$, a rough estimate of the appropriate  $k$ can be obtained by
matching the difference in the $\Tau$ values of the neighboring nodes to the
smear $\sigma$ from eq.~\ref{eq:smear}.  
\begin{equation}
\sigma = \frac{|\Tau_{N+1}|}{\sqrt{k-2}} \left[ \frac{k}{k-1} \right] \simeq |\Tau_{N+1} - \Tau_{N} |  \qquad 
\Rightarrow \qquad \frac{k}{(k-1)\sqrt{k-2}}  \simeq \frac{c}{1+c}.
\label{eq:optimalk}
\end{equation} 
This condition implies that a large value of $k$ will be required when $c$ is
small and a small value of $k$ will be required when $c$ is large. In
particular, $k \simeq 8$ when $c=1$, which will be the parameters we pick for
the applications in the next section.  

To further illustrate the information loss at high values of $k$,
fig.~\ref{fig:optimalk}b shows the sum of activity of the $N$-th and the
$(N+1)$-th nodes for all values of $\tau_o$ between $\Tau_{N}$ and
$\Tau_{N+1}$. For each $k$, the activities are normalized so that the $N$-th
node attains 1 when $\tau_o=\Tau_N$. Focusing on  the case of  $k=100$ in
fig.~\ref{fig:optimalk}b, there is a range of $\tau_o$ values for which the
total activity of the two nodes is very close to zero.  The input is
represented by the $N$-th node when $\tau_o$ is close to $\Tau_N$, and
is represented by the $(N+1)$-th node when $\tau_o$ is close to
$\Tau_{N+1}$, but at intermediate values of $\tau_o$ the input is not
represented by any node.  One way to avoid such information loss is to require
that the total activity of the two nodes should not have a local minimum---in
other words the minimum should be at the boundary, at $\tau_o= \Tau_{N+1}$, as
seen in figure~\ref{fig:optimalk}b for $k=$4, 8 and 12. For $c=1$, it turns
out that there exists a local minimum in the summed activity of the two nodes
only for values of  $k$ greater than 12. For any given $c$, the appropriate value of $k$ that simultaneously minimizes the information redundancy and information loss is the maximum value of $k$ for which a plot similar to fig.~\ref{fig:optimalk}b will not have a local minimum. 
 
In summary, the fuzzy memory system is the set of $\T{}$ column nodes with $\Tau$ values given by eq.~\ref{eq:Tlist}, with the value of $k$ appropriately matched with $c$ to minimize information redundancy and information loss.

\section{Time series forecasting}

We compare the performance of the self-sufficient fuzzy memory to a shift
register in time series forecasting with a few simple illustrations.  Our goal
here is to illustrate the differences between a simple shift register and the
self-sufficient fuzzy memory.  Because our interest is in representation of
the time series and not in the sophistication of the learning algorithm, we
use simple linear regression algorithm to learn and forecast these time
series.

\begin{figure}
\begin{center}
\includegraphics[height=0.6\textheight]{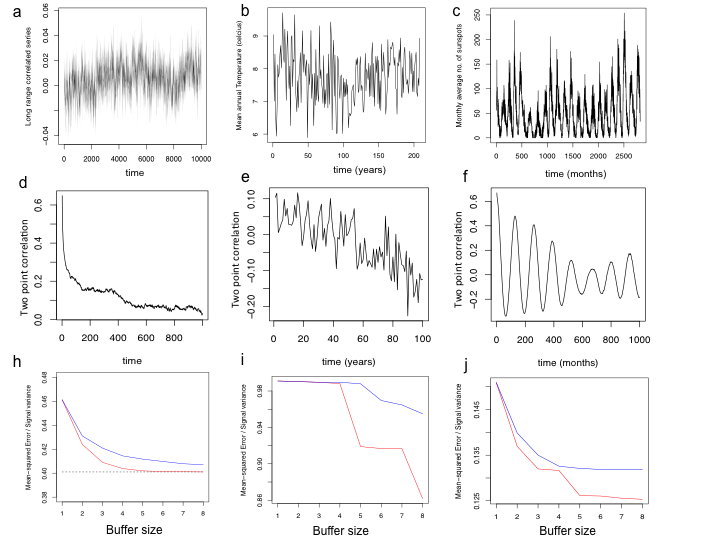}
\end{center}
\caption{Time series forecasting. \textbf{a}.  Simulated time
series with long range correlations based on ARFIMA model with $d=0.4$, and
white noise of standard deviation 0.01. \textbf{b}.  Average
annual temperature of the Earth from the year 1781 to 1988. \textbf{c}.
Monthly average number of sunspots from the year 1749 to 1983.  The top row
gives the values in each series.  The middle row
shows the two point correlation functions. The bottom row shows the error in
forecasting the series using the self-sufficient fuzzy memory (red) and using
the shift register (blue).
 }
\label{fig:Tseries}
\end{figure}

We consider three time series with different properties.  The first was
generated  by fractionally integrating white noise \citep{WageEtal04}  in a
manner similar to that described in section~2.  The second and third time
series were obtained from the online library at http://datamarket.com. The
second time series is the mean annual temperature of the Earth  from the year
1781 to 1988. The third time series is the monthly average number of sunspots
from the year 1749 to 1983 measured from Zurich, Switzerland. These three time
series are plotted in the top row of fig.~\ref{fig:Tseries}. The corresponding
two point correlation  function of each series is plotted in the middle row of
fig.~\ref{fig:Tseries}.  Examination of the two point correlation functions
reveal differences between the series.  The fractionally-integrated noise
series shows long-range correlations falling off like a power law.  The
temperature series shows correlations near zero (but modestly positive) over
short ranges and weak negative correlation over longer times.  The sunspots
data has both strong positive short-range autocorrelation and a longer range
negative correlation, balanced by a periodicity of 130 months corresponding
to the 11 year solar cycle.

\subsection{Learning and forecasting methodology}

Let $\NMAX$ denote the total number of nodes in the memory representation and let $N$ be an
index corresponding to each node ranging from 1 to  $\NMAX$. We shall denote
the value contained in the nodes at any time step $i$ by $B_{i}[N]$.
The time series was sequentially fed into both the shift register and the
self-sufficient fuzzy memory and the representations were evolved appropriately at each time step. The
values in the shift register nodes were shifted downstream at each time step
as discussed section~2. At any instant the shift register held information
from exactly $\NMAX$ time steps in the past. The values in the self-sufficient
fuzzy memory 
were evolved as  described in section~3, with $\Tau$ values  taken to be 1, 2,
4, 8, 16, 32,...$2^{(\NMAX-1)}$, conforming to eq.~\ref{eq:Tlist} with $\Tau_{min}=1$, $c=1$ and $k=8$. 

At each time step $i$, the value from each of the nodes $B_{i}[N]$ was recorded
along with the value of the time series at that time step, denoted by $V_{i}$. We used a
simple linear regression algorithm to extract  the intercept $I$ and the regression coefficients
$R_N$ so that the predicted value of the time series at each time step $P_{i}$  and
the squared error in prediction  $E_{i}$  are 
\begin{equation}
P_{i} =  I + \sum_{N=1}^{\NMAX} R_{N} B_{i}[N], \qquad  \,\, E_{i} = [ P_{i} - V_{i} ]^2 . 
\end{equation}
The regression coefficients were extracted by minimizing the total
squared error $E= \sum_{i} E_{i} $. For this purpose, we used a standard
procedure {\tt lm()} in the open source software  R. 

The accuracy of forecast is inversely related to the total squared error $E$.
To get an absolute measure of accuracy we have to factor out the intrinsic
variability of the time series. In the bottom row of fig.~\ref{fig:Tseries},
we plot the mean of the squared error divided by the intrinsic variance in the
time series $\textrm{Var}[V_{i}]$. This
quantity would range from 0 to 1; the closer it is to zero, the more accurate
the prediction.

\paragraph{Long range correlated series:}
The long range correlated series (fig.~\ref{fig:Tseries}a) is by definition
constructed to yield a two point correlation that decays  as a power law. 
This is evident from its  two point correlation in
fig.~\ref{fig:Tseries}d that is decaying, but always positive. 
Since the value of the series at any time step is
highly correlated with its value at the previous time step, we can expect to
generate a reasonable forecast using a single node that holds the
value from the previous time step. This can be seen from
fig.~\ref{fig:Tseries}h, where the  error in forecast is only 0.45 with a
single node. Adding more nodes reduces the error for both the
shift register and the self-sufficient fuzzy memory. But for a
given number of nodes, the fuzzy memory always has a lower error than the
shift register. This can be seen from fig.~\ref{fig:Tseries}h where the
red curve (corresponding to the fuzzy memory) is below the blue (corresponding
to the shift register). 

Since this series is generated by fractionally integrating white noise, the
mean squared error cannot in principle be lower than the variance of the white
noise used for construction. That is, there is a lower bound for the error
that can be achieved in fig.~\ref{fig:Tseries}h. The dotted line in
fig.~\ref{fig:Tseries}h indicates this bound. Note that the fuzzy memory approaches
this bound with a smaller number of nodes than the shift register.

\paragraph{Temperature series:}
The temperature  series (fig.~\ref{fig:Tseries}b) is much more noisy than
the long range correlated series, and seems structureless. This can
be seen from the small values of its two point correlations in
fig.~\ref{fig:Tseries}e.  This is also reflected in the fact that with a small
number of nodes, the error is very high. Hence it can be concluded that
no reliable short range correlation exist in this series. That is, knowing the
average temperature during a given year does not help much in predicting the average
temperature of the subsequent year.  However, there seems to be a weak 
negative correlation at longer scales that could be exploited in
forecasting. Note from fig.~\ref{fig:Tseries}i that with additional nodes the
 fuzzy memory performs better at forecasting and has a lower error in
forecasting than a shift register. This is because the fuzzy memory can
represent much longer timescales than the shift register of equal size, and
thereby exploit the long range correlations that exist.   

\paragraph{Sunspots series:}
The sunspot series (fig.~\ref{fig:Tseries}c)  is  less
noisy than the other two series considered, and it has an oscillatory
structure of about 130 month periodicity.  It has high short range
correlations, and hence even a single node that holds the value from
the previous time step is sufficient to forecast with an error of 0.15,
as seen in fig.~\ref{fig:Tseries}j. As before, with more 
nodes, the fuzzy memory consistently has a lower error in forecasting than the
shift  register with equal number of nodes. Note that when the number of nodes
is increased from 4 to 8, the shift register does not improve in accuracy
while the fuzzy memory continues to improve in accuracy. 

With a single node, both fuzzy memory and shift register essentially just store the information from the previous time step. Because most of the variance in the
series can be captured by the information in the first node, the difference between the
fuzzy memory and the shift register with additional nodes  is not numerically
overwhelming when viewed in fig.~\ref{fig:Tseries}j.
However, there is a qualitative difference in the properties of the signal
extracted by the two memory systems. In order to successfully learn
the 130 month periodicity, the information about high positive short range correlations
is not sufficient, it is essential to also learn the information about the
negative correlations at longer time scales. From fig.~\ref{fig:Tseries}f,
note that the negative correlations exist at a timescale of 50 to 100 months.
Hence in order to learn this information, these timescales have to be
represented. A shift register with 8 nodes cannot represent
these timescales but the fuzzy memory with 8 nodes can.

\begin{figure}
\begin{center}
\includegraphics[height=0.35\textheight]{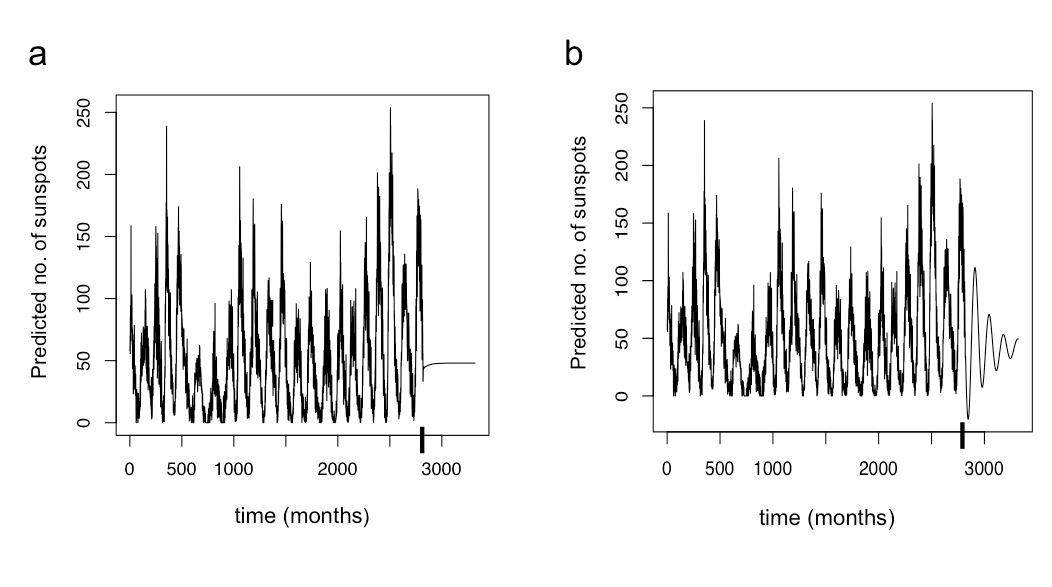}
\end{center}
\caption{Forecasting the distant future. The sunspots time series of length
2820 is extrapolated for 500 time steps in the future using \textbf{(a)} 
a shift register with 8 nodes, and  \textbf{(b)} a self-sufficient fuzzy
memory with 8 nodes.  The solid tick mark on the $x$-axis (at 2820)
corresponds to the point where the original series ends and the predicted
future series begins. 
\label{fig:extend}
}
\end{figure}

To illustrate that it is possible to learn the periodicity using the fuzzy
memory, we forecast the distant future values of the series. In
figure~\ref{fig:extend}, we extend the sunspots series by predicting it for a
future of 500 months. The regression coefficients $R_{N}$ and the intercept $I$ are extracted from
the original series of length 2820. For the next 500 time steps, the
predictions $P_{i}$ are treated as actual values $V_{i}$, and the memory representations are evolved.  
Fig.~\ref{fig:extend}a shows the series generated using  shift register with 8
nodes. 
The solid tick mark on the $x$-axis at 2820 represents the point at which the original series
ends and the predicted future series begins. 
Note that the series forecasted by the shift register immediately settles on the mean value without oscillation.  This is because the time scale at which the oscillations are manifest is not represented by
the shift register with 8 nodes.
Fig.~\ref{fig:extend}b shows the series generated by the fuzzy memory with 8 nodes. Note that the
series predicted by the fuzzy memory continues in an oscillating fashion with
decreasing amplitude for several cycles eventually settling at the mean value.
This is possible because the fuzzy memory represents the signal at a
sufficiently long time scale to capture the negative correlations in the
two-point correlation function.  

Of course, a shift register with many more nodes can capture the long-range
correlations and  predict the periodic oscillations in the signal. However the
number of nodes necessary to describe the oscillatory nature of the signal
needs to be of the order of the periodicity of the oscillation, about 130 in
this case. This would lead to overfitting the data. At least in the case of
the simple linear regression algorithm, the number of  regression coefficients
to be extracted from the data increases with  the number of nodes, and
extracting a large number of regression coefficients from a finite data set
will unquestionably lead to overfitting the data. Hence it would be ideal to
use the least number of nodes required to span the relevant time scale.     

\begin{figure}
\begin{center} 
\begin{tabular}{lclc}
\textbf{a} && \textbf{b}\\
&\includegraphics[height=0.30\textheight]{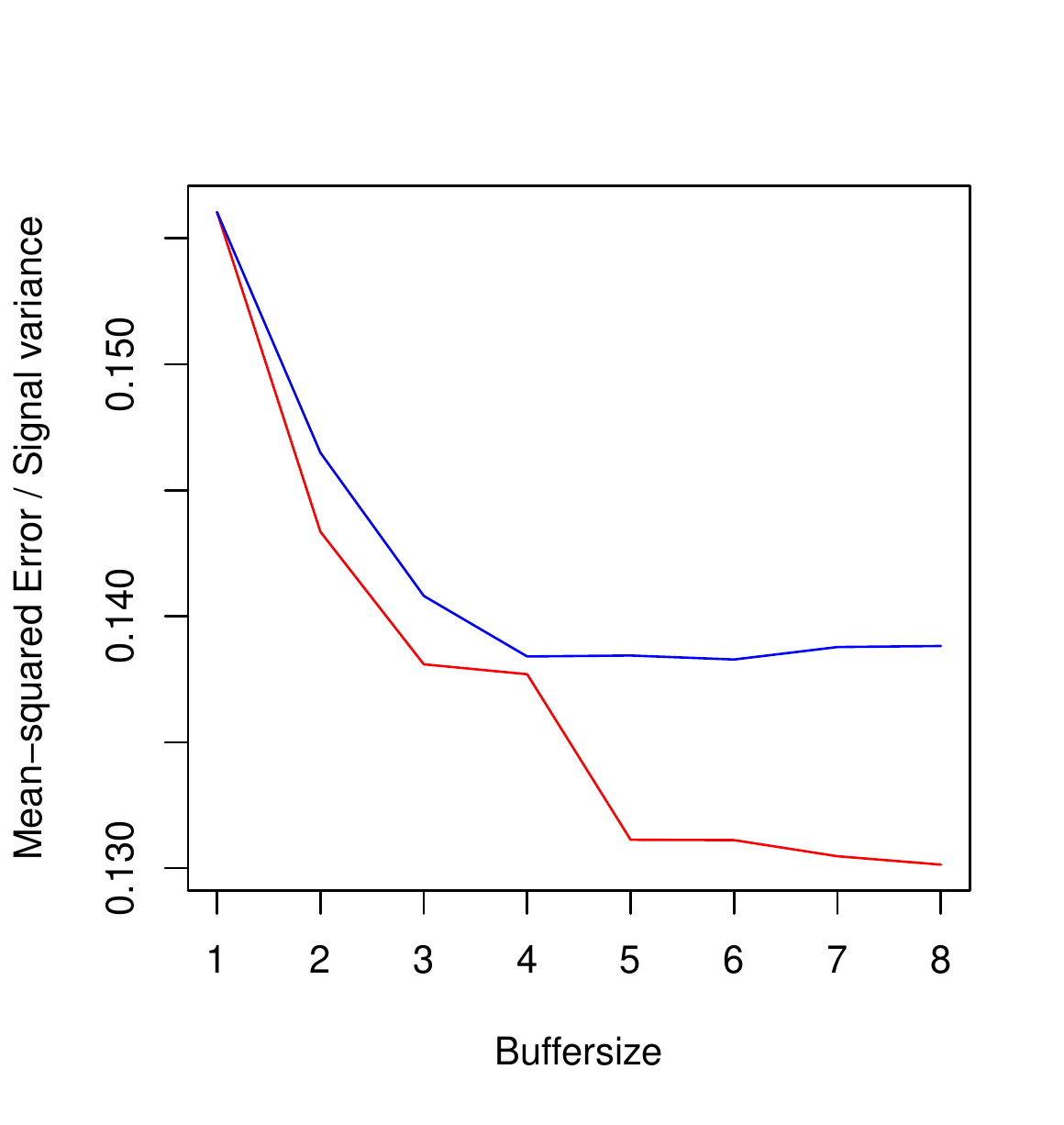} 
&&\includegraphics[height=0.30\textheight]{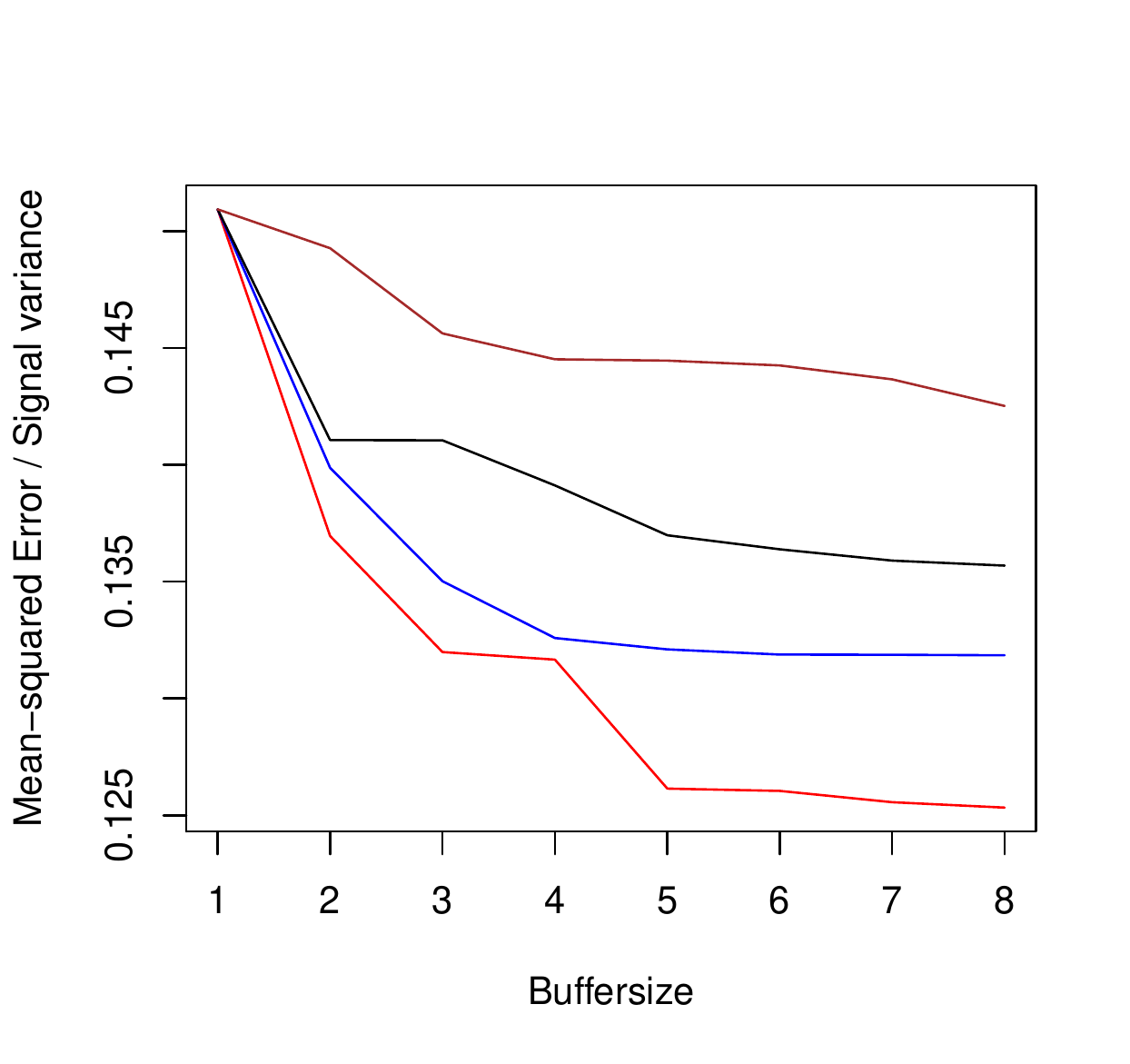}
\end{tabular}
\end{center}
\caption{
\textbf{a.} Testing error in forecasting. The regression coefficients were
extracted from the first half of the sunspot series and testing was performed
on the second half of the series.  Compare to Figure~\ref{fig:Tseries}c,
bottom row. \textbf{b.} Training error of the
self-sufficient fuzzy memory (red), shift register (blue), subsampled shift
register with a  spacing of 5 (black) and subsampled shift register with a
spacing of 10 (brown).  }
\label{fig:test}
\end{figure}

In order to ensure that the extracted regression coefficients has not
overfitted the data, we split the sunspots time series into two halves. We
extracted the regression coefficients by using only the first half for
training and used the second half for testing the predictions generated by
those coefficients. Fig.~\ref{fig:test}a plots this testing
error, and should be compared to the training error plotted in
fig.~\ref{fig:Tseries}j. Other than the noticeable fact that the testing error
is slightly higher than the training error, the shape of the two plots are
very similar for both fuzzy memory (red) and the shift register (blue).  
 
If our goal was to only capture the
oscillatory structure of the sunspot series within a small number of
regression coefficients, then we could subsample from a
lengthy shift register so that information from both positive and negative
correlations can be obtained. 
Although the subsampled shift register contains relatively few nodes, it
cannot self sufficiently evolve in real time; we would need 
the resources associated with the complete shift register in order to evolve
the memory at each moment.     
Subsampling with 8~equidistantly spaced
nodes of the shift register 1,11,21,31,...81, and extracting the corresponding
regression coefficients, it is possible to extend the series to have an
oscillatory structure analogous to fig.~\ref{fig:extend}b. However it turns
out that  the forecasting error for the subsampled shift register is still
higher than the forecasting error from the self-sufficient fuzzy memory. In
fig.~\ref{fig:test}b, the training error is replotted for
the fuzzy memory (red) and shift register
(blue) as in fig.~\ref{fig:Tseries}j, along with the training error for
subsampled shift register with equidistant node spacing of 5 (black) and 10
(brown). Even though the subsampled shift register with a node spacing of 10 extends over a similar temporal range as the fuzzy memory, and captures the oscillatory structure in the data, the fuzzy memory outperforms it with a lower error. The advantage of the fuzzy memory over the subsampled shift register comes from the property of averaging over many previous values at long time scales rather than picking a single noisy value and using that for prediction. This property helps to suppress unreliable fluctuations that could lead to overfitting the data.

\section{Discussion}

The self-sufficient fuzzy memory holds more predictively relevant information
than a shift register with the same number of nodes for long-range correlated
signals, and hence performs better in time series forecasting such signals.
However, learning the relevant statistics from a lengthy time series is not the same as
learning from very few learning trials.  
To learn from very few learning trials, the learner must necessarily make some generalizations based on some built-in assumptions about the environment. Since the fuzzy memory discards information about the precise time of a stimulus presentation, the temporal inaccuracy in memory can help the learner make such a generalization.
Suppose it is useful for a learner to learn the temporal relationship between
two events, say A and B.  Let the statistics of the world be such that B consistently follows A after a delay period, which on each learning trial is chosen from an unknown distribution.  After many learning trials, a learner relying on a 
shift register memory would be able to sample the entire distribution of delays and
learn it precisely.  But real world learners may have to learn much faster.  Because the fuzzy memory system represents the past information in a smeared fashion, a single training sample 
from the distribution will naturally let the learner make a scale-free
temporal generalization about the distribution of delays between A and B.  
The temporal profile of this generalization will not in general match the true
distribution that could be learned after many learning trials, however the fact that it is available after a single learning trial provides a tremendous advantage for natural learners. 

It then seems natural to wonder if human and animal memory resembles the fuzzy memory system. After all, animals have evolved in the
natural world where predicting the imminent future is crucial for survival.
Numerous behavioral findings on animals and humans are consistent with them
having a memory system with scale-free representation of past events
\citep{BalsGall09,GallGibb00}. In human memory studies, the forgetting curve is
usually observed to follow a scale invariant power law function
\citep{DonkNoso12}. When humans are asked to reproduce or discriminate time
intervals, they exhibit a characteristic scale-invariance in the errors they
produce \citep{RakiEtal98,WearLege08}. This is not just a characteristic
feature in humans, but in a wide variety of animals like rats, rabbits and
pigeons \citep{Robe81,Smit68}.  These findings across behavioral tasks and
species suggest that a scale-free memory  is an adaptive response to a world
with structure at many temporal scales.

\subsection{Neural networks with temporal memory}

Let us now consider the fuzzy memory system in the context of neural networks
with temporal memory. It has been realized that  neural networks with generic
recurrent connectivity can have sufficiently rich dynamics to hold temporal
memory of the past. Analogous to how the ripple patterns on a liquid surface
contains information about the past perturbations, the instantaneous state of
the recurrent network holds the memory of the past which can be simply
extracted by training a linear readout layer. Such networks can be implemented
either with analog neurons--echo state networks \citep{Jaeger01}, or with
spiking neurons--liquid state machines \citep{Maassetal01}. They are known to
be non-chaotic and dynamically stable as long as their spectral radius or the
largest eigenvalue of the connectivity matrix has a magnitude less than one.
Abstractly, such networks with fixed recurrent connectivity can be viewed as a
reservoir of nodes and can be efficiently used for computational tasks
involving time varying inputs, including time series prediction
\citep{WyffelsSchrauwen10}.

The timescale of these reservoirs can be tuned up by introducing leaky
integrator neurons in them \citep{JaegerEtal07}. However, a reservoir with
finite nodes cannot have memory from infinite past. In fact the criterion for
dynamical stability of the reservoir is equivalent to requiring a fading
memory \citep{Jaeger02b}. If we define a memory function of the reservoir to be
the precision with which inputs from each past moment can be reconstructed, it
turns out that the net memory, or the area under the memory function over all
past times, is bounded by the number of nodes in the reservoir $\NMAX$. The
exact shape of the memory function will however depend on the connectivity
within reservoir.  For a simple shift register  connectivity, the memory
function is a step function which is 1 up to $\NMAX$ time steps  in the past
and zero beyond $\NMAX$. But for a generic random connectivity the memory
function decays smoothly, sometimes exponentially and sometimes as a power law
depending on the spectral radius \citep{GanguliEtal08}. For linear recurrent
networks,  it turns out that the memory function is analytically tractable at
least in some special cases \citep{WhiteSompolinsky04, HermansSchrauwen09},
while the presence of any nonlinearity  seems to reduce the net
memory\citep{GanguliEtal08}. By increasing the number of nodes in the
reservoir, the net memory can be increased. But unless the network is well
tailored, for e.g. orthogonal network \citep{WhiteSompolinsky04} or a divergent
feed-forward network \citep{GanguliEtal08}, the net memory grows very slowly
and sub-linearly with the number of nodes. Moreover, analysis of trajectories
in the state-space of a randomly connected network suggests that the net
memory will be very low when the connectivity is dense \citep{WallaceEtal13}.   

The self-sufficient fuzzy memory can be viewed as a special case of a
linear reservoir with a specific, tailored connectivity. The $\ti{}$ nodes
effectively have a diagonal connectivity matrix making them leaky integrators,
and the $\Lk$ is the linear readout weights that approximately extracts the
past inputs. For a white noise input signal, the memory function  decays as a
power law with exponent -1, and the net memory grows linearly with the number
of nodes. 
However, as described above, the accuracy of reconstruction of the past is not
the relevant quantity of interest here, it is the predictive information from
the past that is of interest. Scale-free fluctuations in natural world imply
that it isn't necessary to be accurate; in fact sacrificing accuracy in a
scale-free fashion lets us represent predictive information from exponentially long
timescales.    

\subsection{Unaddressed Issues}

Two basic issues essential to real-world machine learning applications have
been ignored in this work for the sake of theoretical simplicity. First is
that we have simply focused on a scalar time varying
signal, while any serious learning problem would involve multidimensional
signals. When unlimited storage resources are available, each dimension can be
separately represented in a lengthy shift register. To conserve storage
resources associated with the time dimension, we could replace the shift
register with the self-sufficient fuzzy memory. This work however does not address the issue
of conserving storage resources by compressing the intrinsic dimensionality of
the signal. In general, most of the information relevant for future prediction
is encoded in few combinations of  the various dimensions of the signal,
called features. Techniques based on information bottleneck method
\citep{CreuSpre08,CreuEtal09} and slow feature analysis \citep{WiskSejn02} can
efficiently extract these features.
The strategy of representing the time series  in a scale invariantly fuzzy fashion could be seen as complementary to these techniques. For instance, slow feature analysis \citep{WiskSejn02} imposes the slowness principle where low-dimensional features that change most slowly are of interest. If the components of a time varying high dimensional signal is represented in a temporally fuzzy fashion rather than in a shift register, then we could potentially extract the slowly varying parts in an online fashion by examining differences in the activities of the largest two $\Tau$ nodes.

The second issue is that we have ignored the learning and prediction mechanisms
while simply focusing on the memory representation. For simplicity we used the
linear regression predictor in section~5. Any serious application should
involve the ability to learn nonlinearities. Support vector machines (SVM) adopt
an elegant strategy of using nonlinear kernel functions to map the input data
to a high dimensional space where linear methods can be used
\citep{Vapn98,MuellerEtal97}. The standard method for training SVMs on time
series prediction requires feeding in the data from a sliding time window, in
other words providing shift registers as input.  It has recently been
suggested that rather than using standard SVM kernels on sliding time window,
if we used recurrent kernel functions corresponding to infinite recurrent
networks, performance can be improved on certain tasks
\citep{HermansSchrauwen11}. This suggests that the gradually fading temporal
memory of the recurrent kernel functions is more effective than the
step-function memory of shift register used in standard SVMs for time series
prediction.  Training SVMs with standard kernel functions along with fuzzy
memory inputs rather than shift register inputs is an alternative strategy for
approaching problems involving signals with long range temporal correlations.   
Moreover, since $\ti{}$ nodes contain all the temporal information needed to construct the fuzzy memory, directly training the SVMs with inputs from $\ti{}$ nodes could also be very fruitful.

Finally, it should be noted that if our aim is to build an autonomous agent we
need both learning and prediction to happen in an online fashion. Many
widely-used machine learning algorithms, e.g.,  SVMs \citep{Vapn98} and deep
learning networks \citep{HintEtal06}, rely on batch processing which requires
the availability of the entire data set prior to learning. Autonomous agents
with limited memory resources cannot adopt such learning strategies. The
learning mechanism cannot rely on information other than what is
instantaneously available in the memory. An online learning algorithm tailored
to act on the fuzzy memory representation could potentially be very useful for
autonomous agents with finite memory resources.

\section{Conclusion}

Signals with long-range temporal correlations are found throughout the natural
world. Such signals present a distinct challenge to machine learners that rely
on a shift-register representation of the time series.  Here we have described
a method for constructing a self-sufficient scale-free representation of
temporal history.  The nodes are chosen in a way that minimizes information
redundancy and information loss while equally distributing them over all time
scales.  Although the temporal accuracy of the signal is sacrificed,
predictively relevant information from exponentially long timescales is
available in the fuzzy memory system when compared to a shift register with
the same number of nodes. This could be an extremely useful way to represent
time series with long-range correlations for use in machine learning
applications. 

\section*{Appendix: Information redundancy across nodes}

\renewcommand\theequation{A\arabic{equation}}
\setcounter{equation}{0}

The information redundancy in the memory representation can be quantified by
deriving expressions for mutual information shared between neighboring nodes.
When the input signal is uncorrelated or has scale-free long range
correlations, it will be shown that the information redundancy is equally
spread over all nodes only when the $\Tau$ values of the nodes are
given by eq.~\ref{eq:Tlist}.  

 Taking $\f{}(\tau)$ to be a stochastic signal and the current moment to be $\tau=0$, the activity of a $\Tau$ node in the $\T{}$ column is (see eq.~\ref{eq:total_T})
\begin{equation}
\T{}(0,\Tau) = \frac{k^{k+1}}{k!} \int_{-\infty}^{0} \frac{1}{|\Tau|} \left( \frac{\tau'}{\Tau} \right)^k e^{-k \left(\frac{\tau'}{\Tau} \right) } \f{}(\tau') d \tau'   .
\end{equation}
The expectation value of this node can be calculated by simply averaging over $\f{}(\tau')$ inside the integral,  which should be a constant if it is generated by a stationary process. By defining $z= \tau'/\Tau$, we find that
the expectation of $\T{}$ is proportional to the expectation of $\f{}$.
\begin{equation}
\EXP{\T{}(0,\Tau)} = \EXP{\f{}} \frac{k^{k+1}}{k!} \int_{0}^{\infty} z^{k} e^{-kz} dz 
\end{equation}
To understand the information redundancy in terms of correlations among the nodes, we calculate the  correlations among the $\T{}$ nodes when $\f{}(\tau)$ is white noise and long-range correlated noise.

\subsection*{White-noise $\f{}(\tau)$}

Let  $\f{}(\tau)$ to be white noise, that is $\EXP{\f{}}=0$ and $\EXP{\f{}(\tau) \f{}(\tau')} \sim \delta(\tau-\tau')$. The variance in the activity of each $\Tau$ node is then given by 
\begin{eqnarray}
\EXP{\T{}^2(0,\Tau)} &=& \left(\frac{k^{k+1}}{k!}\right)^2 \int_{-\infty}^{0}  \int_{-\infty}^{0}  \frac{1}{|\Tau|^2} \left( \frac{\tau}{\Tau} \right)^k e^{-k \left(\frac{\tau}{\Tau} \right) }  \left( \frac{\tau'}{\Tau} \right)^k e^{-k \left(\frac{\tau'}{\Tau} \right) } \EXP{\f{}(\tau) \f{}(\tau')} d \tau d\tau' 
\nonumber \\
&=& \frac{1}{|\Tau|} \left(\frac{k^{k+1}}{k!}\right)^2 \int_{0}^{\infty} z^{2k} e^{-2kz} \,dz
\label{eqwhiteVar}
\end{eqnarray}
As expected, the variance of a large $|\Tau|$ node is small because the activity in this node is constructed by integrating the input function over a large timescale. This induces an artificial temporal correlation in that node's activity which does not exist in the input function. To see this more clearly, we calculate the correlation across time in the activity of one node, at time $\tau$ and $\tau'$. With the definition $\delta = |\tau-\tau'|/|\Tau|$, it turns out that
\begin{equation}
\EXP{\T{}(\tau,\Tau)\ \T{}(\tau',\Tau)}= 
|\Tau|^{-1} \left(\frac{k^{k+1}}{k!}\right)^2
e^{-k\delta} \sum_{r=0}^{k}  \delta^{k-r} \frac{k!}{r!(k-r)!} \int_{0}^{\infty}  z^{k+r} e^{-2kz }  \, dz
\label{eq:tt'cor}
\end{equation}  
Note that this correlation is nonzero for any $\delta>0$, and it decays exponentially for large $\delta$. Hence even a temporally uncorrelated white noise input leads to short range temporal correlations in a $\Tau$ node. It is important to emphasize here that such temporal correlations will not be introduced in a shift register. This is because, in a shift register the functional value of $\f{}$ at each moment is just passed on to the downstream nodes without being integrated, and the temporal autocorrelation in the activity of any node will simply reflect the temporal correlation in the input function.   

Let us now consider the instantaneous correlation in the activity of two
different nodes.  At any instant, the activity of two different nodes in a
shift register will be uncorrelated in response to a white noise input. The
different nodes in a shift register carry completely different information,
making their mutual information zero. But in the $\T{}$ column, since the
information is smeared across different $\Tau$ nodes, the mutual information
shared by different nodes is non-zero. The  instantaneous correlation between
two different nodes $\Tau_1$ and $\Tau_2$ can be calculated to be
\begin{eqnarray}
\EXP{\T{}(0,\Tau_1)\ \T{}(0,\Tau_2)} &=&
 |\Tau_2|^{-1} \left(\frac{k^{k+1}}{k!}\right)^2  \int_{0}^{\infty} z^{2k} (\Tau_1/\Tau_2)^k e^{-kz(1+\Tau_1/\Tau_2) }  \, dz
 \nonumber \\
 & \propto & \frac{(\Tau_1 \Tau_2)^{k}}{(|\Tau_1| +|\Tau_2|)^{2k+1}}
 \label{eq:t1t2}
\end{eqnarray} 
The instantaneous correlation in the activity of the  two nodes $\Tau_1$ and
$\Tau_2$ is a measure of the mutual information represented by them.
Factoring out the individual variances of the two nodes, we have the following
measure for the mutual information.  
\begin{equation}
\mathcal{I}(\Tau_1,\Tau_2) = \frac{ \EXP{\T{}(0,\Tau_1)\ \T{}(0,\Tau_2) } }{ \sqrt{ \EXP{\T{}^2(0,\Tau_1)}   \EXP{\T{}^2(0,\Tau_2)} } }
\propto \left[\frac{\sqrt{\Tau_1/\Tau_2} }{ (1+\Tau_1/\Tau_2) } \right]^{2k+1}
\label{eq:Inf}
\end{equation}
This quantity is high when $\Tau_1/\Tau_2$ is close to 1. That is, the mutual information shared between neighboring nodes will be high when their $\Tau$ values are very close.  

The fact that the mutual information shared by neighboring nodes is
non-vanishing implies that  there is  redundancy in the representation of the
information. If we require the information redundancy to be equally
distributed over all the nodes, then we need the mutual information between
any two neighboring nodes to be a constant. If $\Tau_1$ and $\Tau_2$ are any
two neighboring nodes, then in order for $\mathcal{I}(\Tau_1,\Tau_2)$ to be a
constant,  $\Tau_1 /\Tau_2$ should be a constant.  This can happen only if the
$\Tau$ values of the nodes are arranged in the form given by
eq.~\ref{eq:Tlist}.  

\subsection*{Inputs with long range correlations}

Now consider $\f{}(\tau)$ such that $\EXP{\f{}(\tau) \f{}(\tau')} \sim 1/|\tau - \tau'|^{\alpha}$ for large values of $|\tau-\tau'|$. Reworking the calculations analogous to those leading to eq.~\ref{eq:tt'cor}, we find that the temporal correlation is 
\begin{equation}
\EXP{\T{}(\tau,\Tau)\ \T{}(\tau', \Tau)} = \frac{ |\Tau|^{-\alpha}}{2. 4^k} \left(\frac{k^{k+1}}{k!}\right)^2
\sum_{r=0}^{k} C_r \int_{-\infty}^{\infty}  \frac{|v|^{k-r}}{ | v+\delta |^{\alpha}} e^{-k|v|}  \,  dv  .
\end{equation} 
Here $\delta=|\tau-\tau'|/|\Tau|$ and $C_r =  \frac{k! k+r)!}{r!(k-r)! }
\frac{ 2^{k-r}}{(k)^{k+r+1}} $. The exact value of $C_r$ is unimportant and we only need to note that it is a positive number. 
  
For $\alpha >1$, the above integral diverges at $v=-\delta$, however we are only interested in the case $\alpha<1$. When $\delta$ is very large, the entire contribution to the integral comes from the region $|v| \ll \delta$ and the denominator of the integrand can be approximated as $\delta^{\alpha}$.  In effect, 
\begin{equation}
\EXP{\T{}(\tau,\Tau)\ \T{}(\tau',\Tau)} \sim |\Tau|^{-\alpha} \delta^{-\alpha} = |\tau-\tau'|^{-\alpha} 
\end{equation}
for large $|\tau-\tau'|$. The temporal autocorrelation of the activity of any node should exactly reflect the temporal correlations in the input when $|\tau-\tau'|$ is much larger than the time scale of integration of that node ($\Tau$). As a point of comparison, it is useful to note that any node in a shift register will also exactly reflect the correlations in the input.   

Now consider the instantaneous correlations across different nodes. The instantaneous correlation between two nodes $\Tau_1$ and $\Tau_2$ turns out to be
\begin{equation}
\EXP{\T{}(0,\Tau_1)\ \T{}(0,\Tau_2)}=  |\Tau_2|^{-\alpha} \left(\frac{k^{k+1}}{k!}\right)^2
\sum_{r=0}^{k} X_r \frac{\beta^{k-r}(1+ \beta^{r-\alpha +1})}{(1+ \beta)^{2k-r+1}}  .
\label{eq:instcorr}
\end{equation}
Here $\beta = |\Tau_1|/|\Tau_2|$ and each $X_r$ is a positive coefficient. By always choosing $|\Tau_2| \ge |\Tau_1|$, we note the two limiting cases of interest, when $\beta \ll 1$ and when $\beta \simeq 1$. When $\beta \ll 1$, the $r=k$ term in the summation of the above equation yields the leading term, and the correlation is simply proportional to $|\Tau_2|^{-\alpha}$, which is approximately equal to $|\Tau_2 - \Tau_1|^{-\alpha}$. In this limit where $\Tau_2| \gg |\Tau_1|$, the correlation between the two nodes behaves like the correlation between two shift register
nodes. When $\beta \simeq 1$, note from eq.~\ref{eq:instcorr} that the correlation will still be proportional to $|\Tau_2|^{-\alpha}$. Now if $\Tau_1$ and $\Tau_2$ are neighboring nodes with close enough values, we can evaluate the mutual information between them to be
\begin{equation}
\mathcal{I}(\Tau_1,\Tau_2) = \frac{\EXP{\T{}(0,\Tau_1)\ \T{}(0,\Tau_2)}}{\sqrt{\EXP{\T{}^2(0,\Tau_1)}\EXP{\T{}^2(0,\Tau_2)} }}
\propto  |\Tau_2 / \Tau_1 |^{-\alpha/2} .
\end{equation}

Reiterating our requirement from before that the mutual information shared by neighboring nodes at all scales should be the same, we are once again led to choose $\Tau_2/\Tau_1$ to be a constant which is possible only when the $\Tau$ values of the nodes are given by eq.~\ref{eq:Tlist}.

\section*{Acknowledgements}
The authors  acknowledge support from National Science Foundation grant,  NSF BCS-1058937, and Air Force Office of Scientific Research grant AFOSR FA9550-12-1-0369. 

\section*{References}


\end{document}